\RequirePackage{fix-cm}
\documentclass[smallcondensed]{svjour3}       
\smartqed  
\usepackage{graphicx}
\usepackage{amsmath}
\usepackage{listings}
\usepackage{textcomp}
\usepackage{color}
\usepackage{ amssymb }
\usepackage{physics}
\usepackage{siunitx}
\usepackage{ stmaryrd }
\usepackage{subcaption}
\usepackage{gensymb}
\usepackage{caption}
\usepackage{multicol}
\usepackage{mathtools}
\usepackage{tabularx} 
\usepackage{multirow}
\usepackage{caption}
\usepackage{fixltx2e}
\usepackage{academicons}
\usepackage{xcolor}
\usepackage{svg}
\usepackage{fontawesome5}
\usepackage{geometry}
\usepackage{hyperref}
\usepackage{siunitx}
\usepackage{calrsfs}
\usepackage{mathrsfs}
\usepackage{amsfonts}
\usepackage[cal = cm]{mathalpha}
\usepackage{cite}
\usepackage{mathptmx}      
\usepackage{latexsym}

\pdfoutput=1
\begin{document}
\title{Unveil the unseen: exploit information hidden in noise 
}


\author{Bahdan Zviazhynski      \and
        Gareth Conduit
}


\institute{Bahdan Zviazhynski \at
              Theory of Condensed Matter Group, Cavendish Laboratory,
University of Cambridge, J. J. Thomson Avenue, Cambridge,
CB3 0HE, UK \\
              \email{bz267@cam.ac.uk}           
           \and
           Gareth Conduit \at
Theory of Condensed Matter Group, Cavendish Laboratory,
University of Cambridge, J. J. Thomson Avenue, Cambridge,
CB3 0HE, UK \\
              \email{gjc29@cam.ac.uk}  
}


\maketitle

\begin{abstract}
Noise and uncertainty are usually the enemy of machine learning, noise in training data leads to uncertainty and inaccuracy in the predictions. However, we develop a machine learning architecture that extracts crucial information out of the noise itself to improve the predictions. The phenomenology computes and then utilizes uncertainty in one target variable to predict a second target variable. We apply this formalism to PbZr$_{0.7}$Sn$_{0.3}$O$_{3}$ crystal, using the uncertainty in dielectric constant to extrapolate heat capacity, correctly predicting a phase transition that otherwise cannot be extrapolated. For the second example -- single-particle diffraction of droplets -- we utilize the particle count together with its uncertainty to extrapolate the ground truth diffraction amplitude, delivering better predictions than when we utilize only the particle count. Our generic formalism enables the exploitation of uncertainty in machine learning, which has a broad range of applications in the physical sciences and beyond.
\keywords{machine learning, uncertainty, extrapolation, case studies}
\end{abstract}
\section{Introduction}
Throughout the human history, scientific discoveries heralded each new epoch, including the stone, bronze, and iron ages. However, discovering new phenomena is not the only challenge: utilizing the freshly obtained knowledge for real-world applications is crucial. With the availability of computers and large amounts of experimental/computational data nowadays \cite{ref42,ref43,ref44,ref48}, machine learning \cite{ref37,ref38,ref39,ref40,ref41} has proven an effective tool for this purpose.

Machine learning is a class of methods that start from existing data to train a model and then predict the quantities of interest useful for a given application. For example, machine learning can predict many properties of a putative material \cite{ref35,ref36,ref23,ref24,ref25,ref28,ref27,ref26,ref59}, and moreover can understand the uncertainty in those predictions. This uncertainty can be used to design the material that is most likely to satisfy the set target criteria \cite{ref1,ref8,ref2}, avoiding the typical expensive and time-consuming cycles of trial and improvement experiments. Furthermore, the uncertainty is useful for accelerating materials discovery by guiding where new experiments should be performed in the materials space \cite{ref18,ref29,ref19}, and also for the identification of outliers and erroneous entries in materials databases \cite{ref45}.

While uncertainty is crucial for focusing on the most viable candidates for a given application, uncertainty itself could be a useful value for property prediction. This strategy is motivated by Wilson's Renormalization Group theory \cite{ref3}, in which fluctuations on all scales determine the macroscopic state of the system. An example is the liquid-vapor transition, in which near the critical temperature water droplets and vapor intermix over all length scales, leading to critical opalescence \cite{ref20}. This means there will be dependences between moments of probability distributions of different quantities, e.g.\,\,uncertainty in one quantity could contain information about the expected value of another quantity. Moreover, combining the expected value and uncertainty in one quantity could increase the amount of information about another quantity, improving the quality of its predictions. An example is shot noise \cite{ref49}, for which the expected value is equal to the square of its uncertainty. Combining both expected value and uncertainty would double the amount of information about the second quantity. 

In this paper, we first review the machine learning methods in the literature and set up the formalism to extract information from uncertainty in Section \ref{s2}. We then validate the formalism on paradigmatic datasets in Section \ref{s3} and apply it to the real-world physical examples -- PbZr$_{0.7}$Sn$_{0.3}$O$_{3}$ crystal phase transitions and single-particle diffraction of droplets -- in Section \ref{s4}, predicting the quantity of interest by extrapolation in both cases. Finally, we discuss broader applications of the generic methodology in Section \ref{s5}. 

\section{Methodology}
\label{s2}
We build the individual components for the machine learning methodology before compiling them into a tool to extract information from noise. First, in Section \ref{s21} we describe the underlying vanilla random forest machine learning method. In Section \ref{s22} we outline how the machine learning algorithm estimates the uncertainty in its predictions. Following this, in Section \ref{s23} we address how to handle missing data using an intermediate target variable before finally putting all three components together in Section \ref{s24} to extract information from uncertainty.  
\subsection{Machine learning}
\label{s21}
Machine learning algorithms are trained on existing data to make predictions of target variables for new data entries. A few examples of widely used machine learning algorithms are \textit{k}-means clustering \cite{ref5}, neural network \cite{ref6} and Gaussian processes \cite{ref9}. In the current work we use random forest, implemented in Scikit-learn Python package \cite{ref13}, since it is computationally cheap and robust against overfitting.
 
Random forest is a collection of independent identical regression trees \cite{ref46}. During the training phase, each tree learns the rules for mapping the input variables to target variables. The geometry of regression trees in a random forest, and therefore accuracy of predictions, is affected by the hyperparameters of the random forest. In order to achieve the best accuracy of predictions, we tune $min\_samples\_leaf$ hyperparameter, which is the minimum number of datapoints in each leaf of a regression tree. 

A robust method of assessing the accuracy of a model during hyperparameter tuning, applicable to any machine learning algorithm, is $k$-fold cross-validation. In this method, of the $k$ equally sized subsamples of training data, each one is retained as the validation data for testing the model trained on the remaining subsamples. The process is repeated $k$ times to obtain the average $R^2$, coefficient of determination, on validation, which is to be maximized. 
\subsection{Uncertainty from machine learning}
\label{s22}
Uncertainty can be extracted from machine learning by bootstrapping \cite{ref5}, a process in which for a dataset with $N$ entries, new subsets are generated from it by sampling $N$ entries randomly with replacement. Each subset is used to train one regression tree in the random forest. The compound predictions of the constituent trees are averaged to give the random forest prediction, and their standard deviation is the uncertainty in this prediction.
\subsection{Use of intermediate target variable to handle missing data}
\label{s23}
\begin{figure}
\centering
\includegraphics[width=0.78\linewidth]{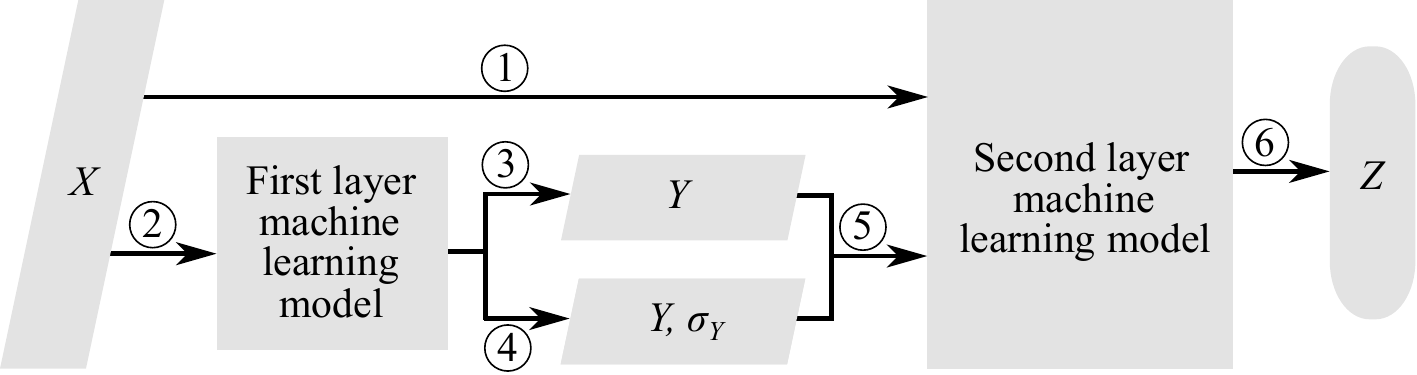}
\caption{Flowchart for the multilayer regressor. Flow 1 takes $X$ directly to the second machine learning model; Flow 2 takes $X$ to the first machine learning model; Flows 3 and 4 predict $Y$ and $Y, \sigma_Y$ respectively; Flow 5 takes $Y$ and $\sigma_Y$ to the second layer; Flow 6 outputs $Z$.}
\label{fig3}
\end{figure}

Data is often missing or even occasionally erroneous in databases, which limits the information available to train the model and make predictions. However, machine learning algorithms can learn and exploit correlations between various properties to help impute the missing values and train a better model or make more accurate predictions. We can visualize the flow of information through the machine learning algorithm with the flowchart in Fig. \ref{fig3}. A standard machine learning approach will follow Flows 1 \& 6 to use only the input feature $X$ to predict the target variable $Z$. Instead, we can train the first machine learning model on $X$ (Flow 2) to predict an intermediate target variable $Y$ (Flow 3), which is correlated with the target variable $Z$. We then train the second machine learning model on $Y$ (Flow 5) to predict $Z$ (Flow 6).

\begin{figure}
\centering
\includegraphics[width=1\linewidth]{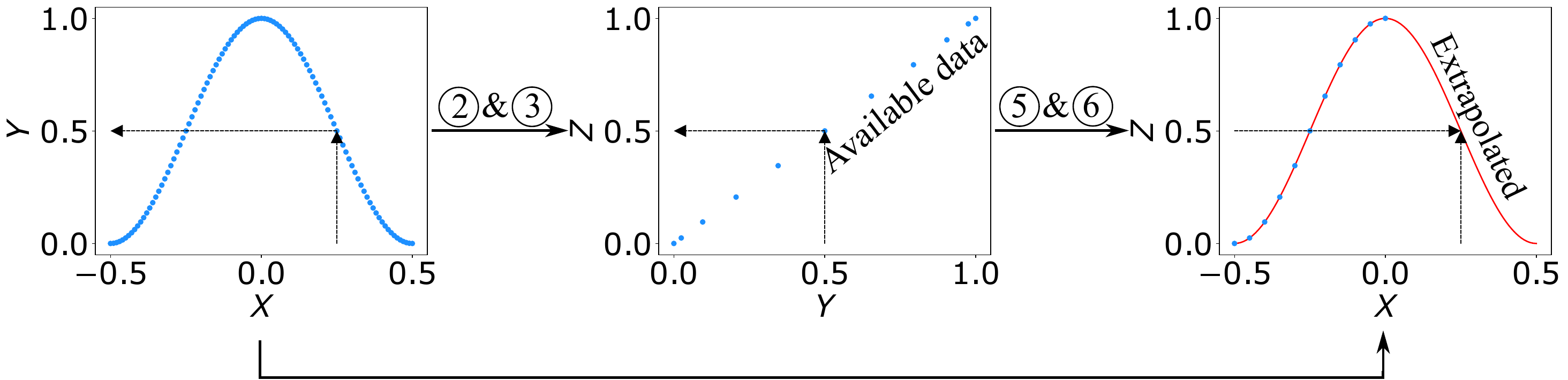}
\caption{The three plots applied sequentially utilize variable $Y$ to extrapolate $Z$ on $X$. In the rightmost figure $Z$-values are missing for $X > 0$. Blue points are training data, red curve is extrapolated values. The arrows represent Flows 2 \& 3 and 5 \& 6 from Fig. \ref{fig3}}
\label{fig2}
\end{figure}

Using the intermediate variable $Y$ is particularly useful if we want to extrapolate $Z$ for $X$ outside the training range, e.g. when we have data available for $Z$ at $X < 0$ but no data for $Z$ at $X > 0$, as in the third plot in Fig. \ref{fig2}. At both $X < 0$ and $X > 0$, data is available for $Y$, as illustrated in the $X-Y$ plot in Fig. \ref{fig2}. Starting by learning the $X-Y$ relationship, and then exploiting the correlation between $Y$ and $Z$, we can improve $Z$-predictions. In other words, two interpolations $X \rightarrow Y$ (Flow 3) and $Y \rightarrow Z$ (Flow 6) would give an extrapolation $X \rightarrow Z$. This situation commonly arises if $Z$ is more expensive to measure than $Y$ and/or more data is available for $Y$ than for $Z$. This strategy has been successfully applied to materials and drugs design to exploit relationships between various properties \cite{ref45,ref52,ref55,ref57}.
\subsection{Use of uncertainty as an input}
\label{s24}
With the three components -- machine learning, estimation of uncertainty, and handling of missing data -- in place, we are well-positioned to develop the overarching framework to use uncertainty as an input for machine learning. In this methodology we will utilize uncertainty in one target variable for extrapolation of another final target variable.

The simplest way to use the uncertainty in target variable $Y$ for extrapolation of another target variable $Z$ is to use a multilayer regressor (Fig. \ref{fig3}). We first train one random forest (first layer regressor) on $X$ (Flow 2) to predict $Y$ and its uncertainty $\sigma_Y$ (Flow 4); then train another random forest (second layer regressor) on $X$ (Flow 1), $Y$ and $\sigma_Y$ (Flow 5) to predict $Z$ (Flow 6). Both first and second layer regressors have identical hyperparameters.

\section{Algorithm validation}
\label{s3}
Having implemented the machine learning algorithm, we now need to validate its performance. Firstly, in Section \ref{s31}, we test the ability to accurately predict uncertainty. Secondly, in Section \ref{s32}, we confirm the ability to use the prediction of one target variable to predict another target variable. Thirdly, in Section \ref{s33}, we validate the core functionality by estimating the uncertainty in one variable and using it to predict the second variable. Finally, in Section \ref{s34}, we validate the ability of the algorithm to use the combination of one variable and its uncertainty to predict another variable.
\subsection{Uncertainty evaluation}
\label{s31}
\begin{figure}
\centering
\begin{subfigure}{.485\textwidth}
  \includegraphics[width=\linewidth]{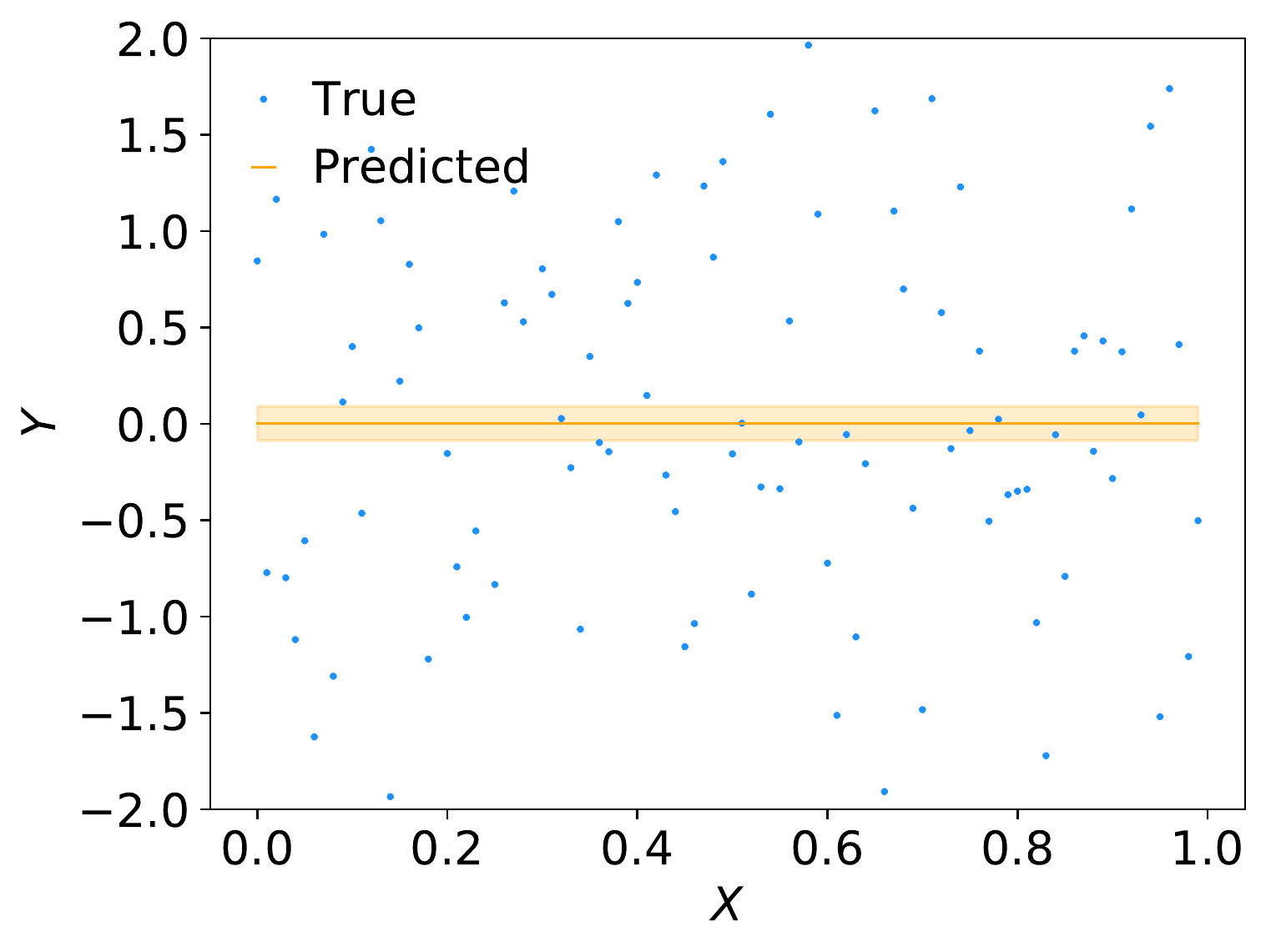}
  \caption{}
  \label{fig4:sub1}
\end{subfigure}%
\begin{subfigure}{.48\textwidth}
  \includegraphics[width=\linewidth]{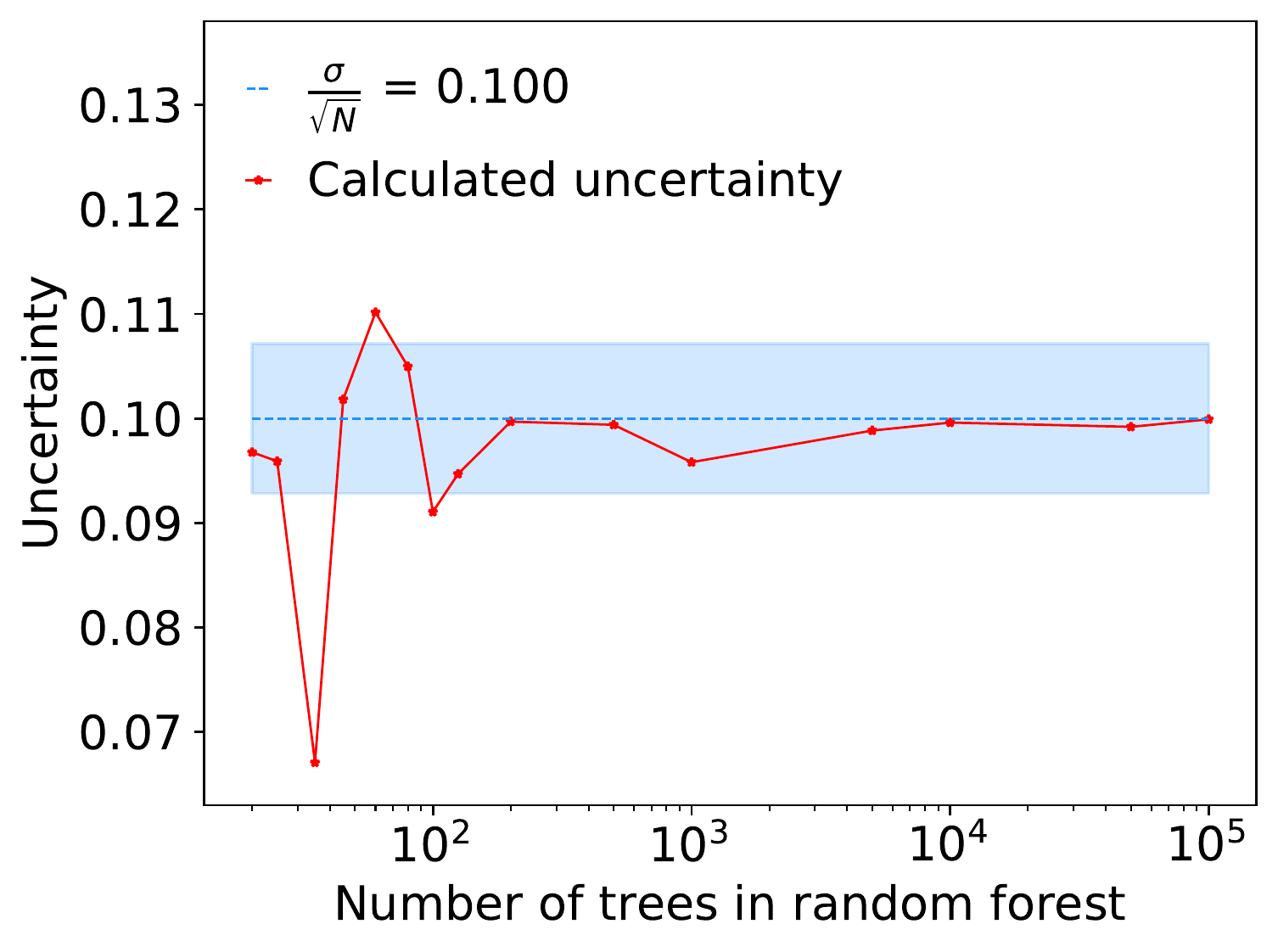}
  \caption{}
  \label{fig4:sub2}
\end{subfigure}
\caption{Random forest for Gaussian white noise: (a) predictions (orange, error region shaded), (b) calculated uncertainty (red) convergence to the true value (blue, error region shaded) as the number of trees is increased.}
\label{fig4}
\end{figure}

Understanding the uncertainty is central to this study, so first we confirm that our machine learning method of choice -- random forest -- gives a good estimate of the uncertainty in its predictions. We adopt a dataset comprising $N = 100$ entries (Fig. \ref{fig4:sub1}), where $0 < X < 1$ and $Y$ is normally distributed white noise $\sim \mathcal{N} (\mu, \sigma^2)$ with the mean $\mu = 0$ and the variance $\sigma^2 = 1^2$. The hyperparameters that maximize the 5-fold cross-validation $R^2$ in $Y$-predictions were found: $min\_samples\_leaf = 100$, leading to each tree averaging over all of the training data, giving a constant prediction equal to the dataset mean, with uncertainty of 0.091 (Fig. \ref{fig4:sub1}). This compares favorably to the analytical estimate of the uncertainty in mean, $\frac{\sigma}{\sqrt{N}} = 0.100 \pm 0.007$ (Fig. \ref{fig4:sub2}, blue dotted line, error region shaded) \cite{ref30}.

\begin{figure}
\centering
\begin{subfigure}{.48\textwidth}
  \includegraphics[width=\linewidth]{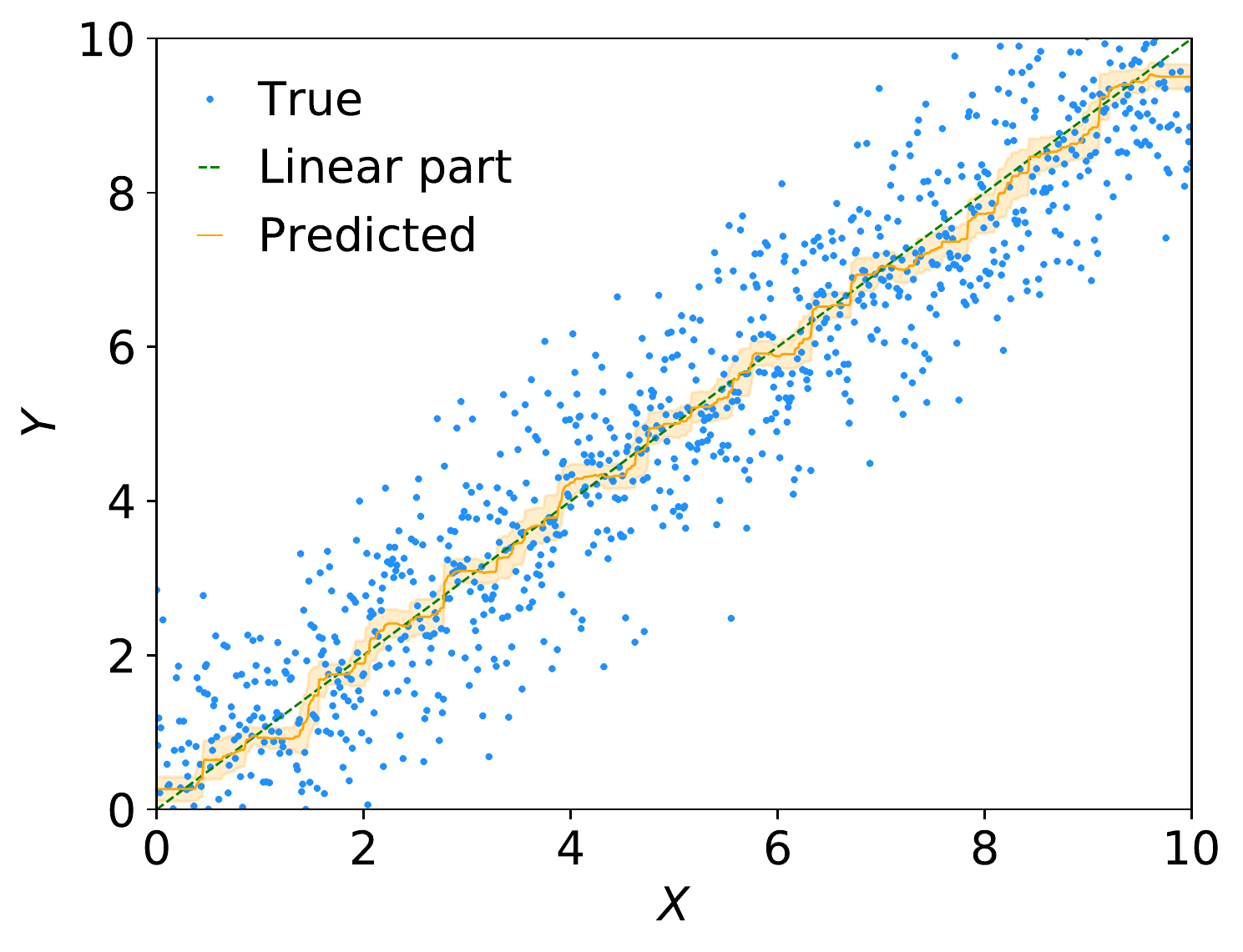}
  \caption{}
  \label{fig5:sub1}
\end{subfigure}%
\begin{subfigure}{.49\textwidth}
  \includegraphics[width=\linewidth]{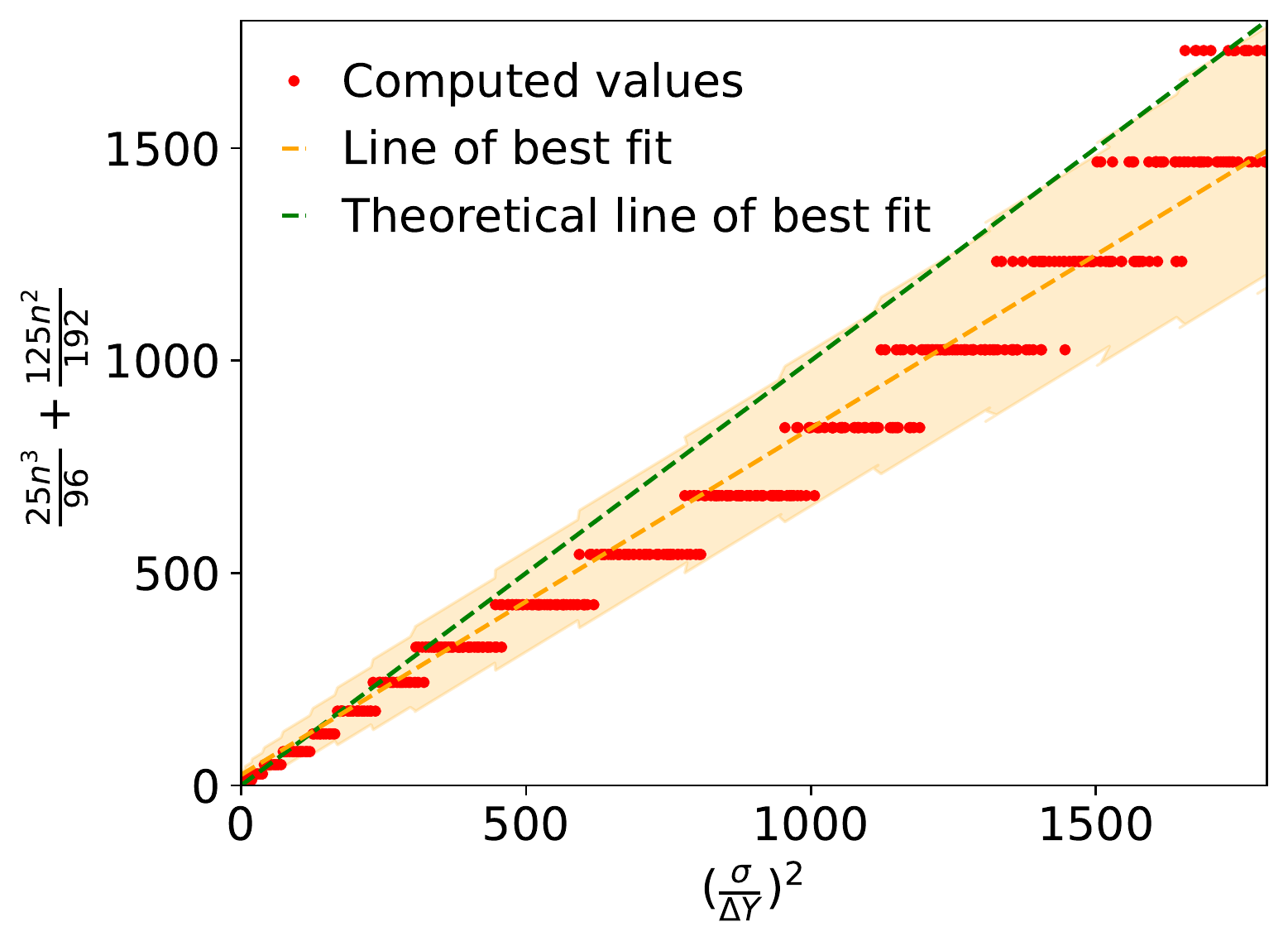}
  \caption{}
  \label{fig5:sub2}
\end{subfigure}
\caption{Random forest model for a linear function with Gaussian white noise: (a) predictions (orange, error region shaded), (b) plot of $\frac{25n^3}{96} + \frac{125n^2}{192}$ against $(\frac{\sigma}{\Delta Y})^2$, red dots are computed values, orange line is the straight line of best fit, blue line is the theoretical line of best fit (slope of 1, through the origin).}
\label{fig5}
\end{figure}

We next check the number of trees, that is number of parallel models trained on different replicas of the data, required to give good estimates of the uncertainty. In Fig. \ref{fig4:sub2} we see that uncertainty estimates using 125 or more trees are all within the error region of $\frac{\sigma}{\sqrt{N}} = 0.100 \pm 0.007$. Therefore, we adopt 125 trees for the remainder of this study.
The choice of $min\_samples\_leaf$ was instrumental to enable the averaging over noise and to give valid uncertainty predictions. Therefore, in order to further investigate the noise averaging lengthscale of random forest, we study a dataset with $0 < X < 10$ and $Y \sim \mathcal{N}(X, 1^2)$ (Fig. \ref{fig5:sub1}). The value of $min\_samples\_leaf$ that minimized the cross-validation MSE in $Y$-predictions was 25, leading to steps in the predictions, which can be seen in Fig. \ref{fig5:sub1}. Should the random forest average over a higher number of adjacent datapoints, the contribution to the MSE from white noise decreases. However, at the same time, the model underfits the underlying linear function even more. This means that for a given noisy dataset there must be an optimal minimum number of datapoints in the tree leaf.

For a more rigorous analysis of the averaging lengthscale, we consider $Y \sim \mathcal{N}(X, \sigma^2)$, and suppose that $n = \textit{min}\_\textit{samples}\_\textit{leaf}$. With points spaced at average increments $\Delta Y$ on the $Y$-axis, $n\Delta Y$ is the increase in the underlying linear function across the leaf. This leads to the average contribution to the $k$-fold cross-validation mean squared error due to underfitting being $\sim \sqrt{\frac{n^2\Delta Y^2}{12} + \frac{n k^2\Delta Y^2}{12(k-1)}}$, where $k\ll n$. On the other hand, the contribution to the mean squared error from the noise in the prediction is $\frac{\sigma}{\sqrt{n\frac{k-1}{k}}}$. The two contributions add in quadrature to give the total squared error, which when minimized with respect to $n$ gives: 
\begin{equation}
\frac{n^3 k^2 \Delta Y^2}{6(k-1)^2} + \frac{n^2 k^3\Delta Y^2}{12(k-1)^2} = \sigma^2
\label{eqn2}
\end{equation}
Performing numerical experiments, we found the optimal $n$ for several values of $\sigma$ using machine learning hyperparameter optimization with 5-fold cross-validation. The values of $\frac{25n^3}{96} + \frac{125n^2}{192}$ -- left-hand side of Equation \ref{eqn2} with $k = 5$ -- plotted against $(\frac{\sigma}{\Delta Y})^2$ can be seen in Fig. \ref{fig5:sub2} as red dots. The plot has a series of plateaus, since $n$ is an integer. A straight line fitted on the red dots (orange line) is within one standard deviation of the theoretical line of best fit (slope of 1, through the origin). This confirms that random forest coupled with hyperparameter optimization gives valid uncertainty predictions, allowing uncertainty to be used as a dependable input for machine learning to predict other quantities later in the paper.
\subsection{Extrapolation using intermediate target variable}
\label{s32}
\begin{figure}
\centering
\begin{minipage}{.49\textwidth}
  \captionof{figure}{
Prediction of $Z$ using $Y$ for extrapolation with one period in training data. (a) $Y$ vs $X$ on training set (blue), (b) $Z$ vs $Y$ on training set, (c) $Z$-predictions given $X$, using $Y$ (red, error region shaded, $R^2 = 0.9995$ on validation) and without $Y$ (orange, error region shaded, $R^2 = -1.75$ on validation). The grey shaded area is the validation set.\vspace{1cm}}
  \label{fig7}
\end{minipage}
\hfill
\begin{subfigure}{.49\textwidth}
  \includegraphics[width=\linewidth]{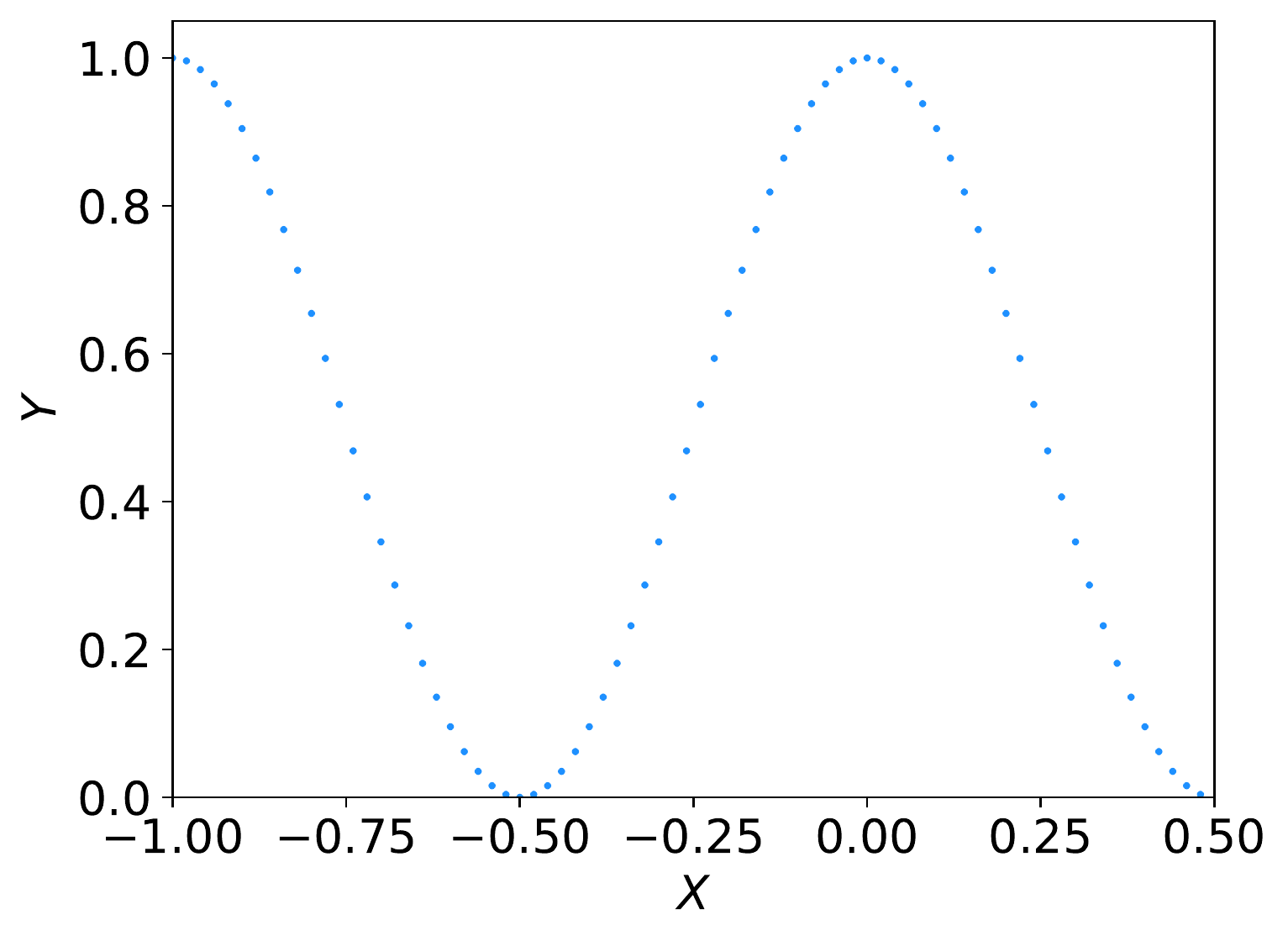}
  \caption{}
  \label{fig7:sub1}
\end{subfigure}\\
\begin{subfigure}{.47\textwidth}
  \includegraphics[width=\linewidth]{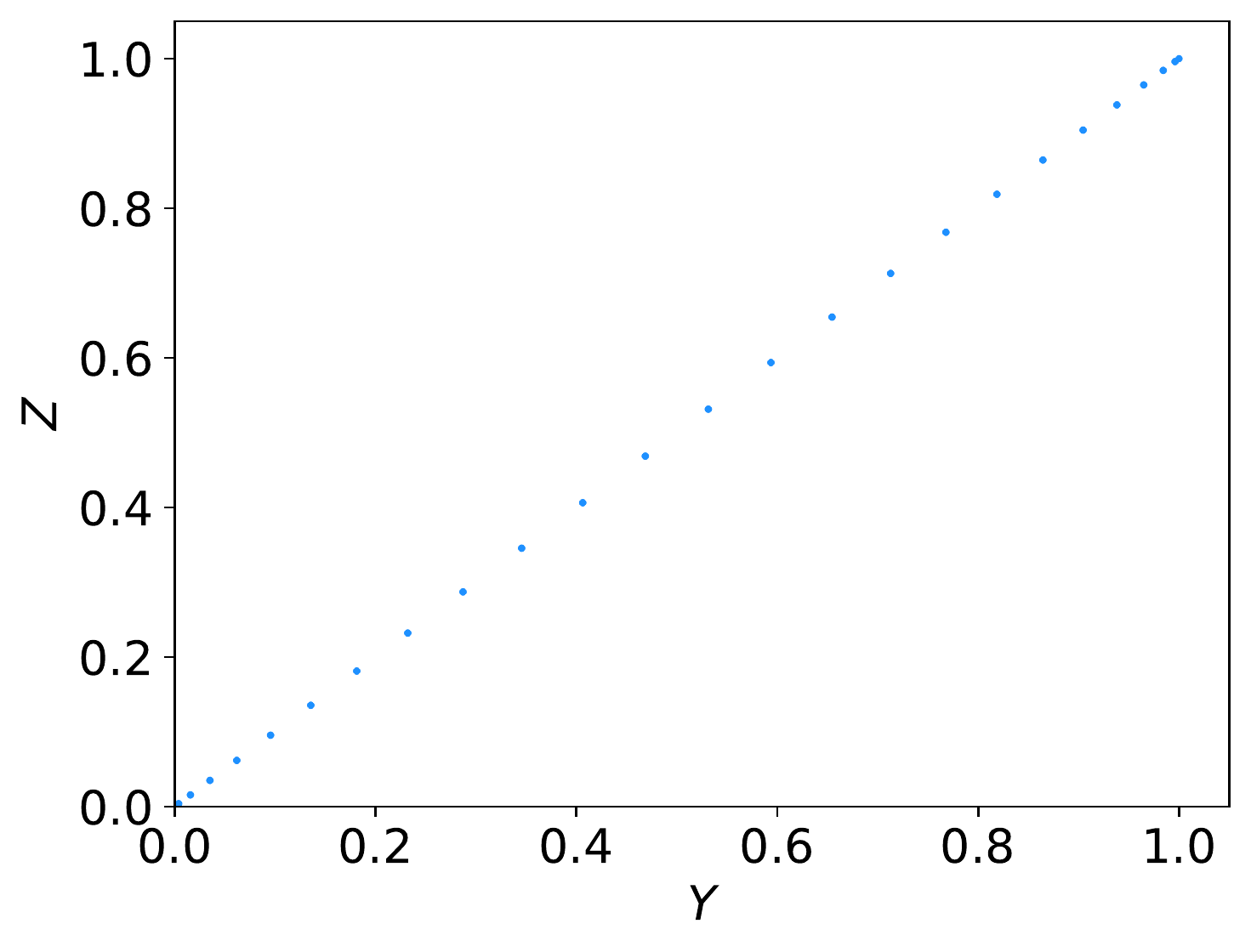}
  \caption{}
  \label{fig7:sub2}
\end{subfigure}
\hfill
\begin{subfigure}{.49\textwidth}
  \includegraphics[width=\linewidth]{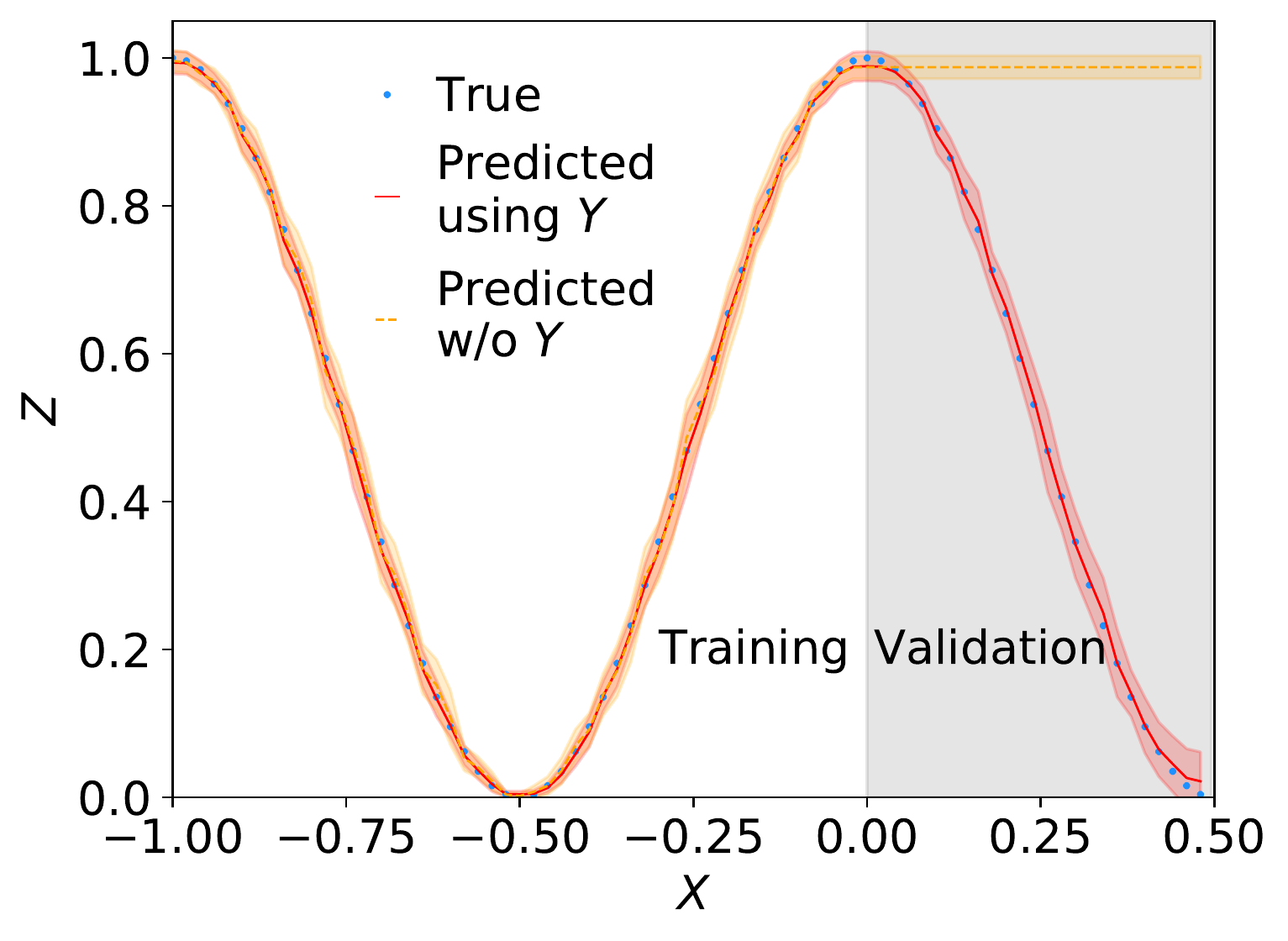}
  \caption{}
  \label{fig7:sub3}
\end{subfigure}
\end{figure}

Having validated that a random forest model delivers reliable estimates of the uncertainty in its predictions, we turn to confirm the capability of the multilayer regressor to utilize target variables for extrapolation. This requires a dataset that comprises three columns: one feature column $X$, an intermediate target column $Y(X)=$ cos$^{2}(\pi X)$ (containing information to guide extrapolation), and final target column $Z(X) = Y(X)=$ cos$^{2}(\pi X)$. In the training set at $-1 < X < 0$, both $Y(X)$ and $Z(X)$ columns have data. At $0 < X < 0.5$, the $Y(X)$ column has data but the $Z(X)$ column is blank -- the missing data that we seek to extrapolate for validation. The data can be seen in Fig. \ref{fig7}, and the region with missing $Z$ is shaded in grey in Fig. \ref{fig7:sub3}, whereas all the data with a white background is present.

First, working on the training set at $X < 0$, we find the hyperparameters that maximize $R^2$ in $Z$-predictions calculated following the blocking cross-validation \cite{ref51}. In this method, of the three equally sized subsamples of the training set split along $X$-axis, the two outermost subsamples are retained as validation data for testing the model trained on the remaining subsamples. The model with the tuned hyperparameters, which give the highest $R^2$ on blocking cross-validation, was trained and then used to predict $Z$ at $X > 0$. The predictions of $Z$ at $X > 0$ were then compared against the true values. There are two strategies to predict $Z(X)$: firstly $X \rightarrow Z$ and secondly $X \rightarrow Y \rightarrow Z$. In Fig. \ref{fig7}, we have a full period of $Z(X)$ ($-1 < X < 0$) in the training set, so it is more straightforward for the random forest model to learn the monotonic $Z = Y$ (Fig. \ref{fig7:sub2}) rather than the oscillating $Z(X)$ (Fig. \ref{fig7:sub3}). Therefore, predictions of $Z$ mostly follow the $X \rightarrow Y \rightarrow Z$ strategy. This leads to better predictions of $Z$ at $X > 0$ with $R^2 = 0.9995$ on validation (Fig. \ref{fig7:sub3}). Moreover, this is confirmed through the random forest feature importances of $X$ and $Y$ for predicting $Z$ being 0.003 and 0.997 respectively.

Having seen the good performance of machine learning algorithm to circumvent missing data to extrapolate $Z(X)$, we also confirm that shifting or scaling $Z(X)$ makes no difference to the accuracy, which is expected as random forest is both shift and scale-invariant.

\subsection{Extrapolation using uncertainty}
\label{s33}
We have shown that the multilayer regressor can accurately evaluate the uncertainty and utilize target variables for extrapolation. We now juxtapose these capabilities and validate the algorithm's ability to utilize uncertainty in one variable to extrapolate another variable. We construct a paradigmatic dataset with $X$, $Y$, and $Z$ columns, where $X$ is the feature column, $Y \sim \mathcal{N}(0, \vert Z(X)\vert^2)$, i.e.\,\,the noise is equal to $Z(X)$ and hence $\sigma_Y \propto Z$. Therefore, the second machine learning model in Fig. \ref{fig3} can learn $X \rightarrow \sigma_Y \rightarrow Z$ (Flows 2, 4, 5 \& 6) more easily than $X \rightarrow Z$ (Flows 1 \& 6), allowing extrapolation beyond the training range of $X$. This is analogous to our study of missing data, simply following Flow 4 rather than Flow 3 in the flowchart in Fig. \ref{fig3}. Therefore, to cement the analogy we adopt the same target function $Z(X)=$ cos$^2(\pi X)$ as before, with $Z$-values missing for $X>0$ for validation.

\begin{figure}[h]
\begin{minipage}{.47\textwidth}
  \captionof{figure}{
Predictions from a random forest trained on one period of Gaussian white noise of periodic amplitude: (a) $Y$ (orange, error region shaded) and $\sigma_Y$ (red) predictions, (b) $Z$ vs $\sigma_Y$ on training set, (c) $Z$-predictions given $X$, using $\sigma_Y$ (red, error region shaded, $R^2 = 0.89$ on validation) and without $\sigma_Y$ (orange, error region shaded). The grey shaded area is the validation set. \vspace{0.1cm}}
  \label{fig8}
\end{minipage}
\hfill
\begin{subfigure}{.52\textwidth}
  \includegraphics[width=\linewidth]{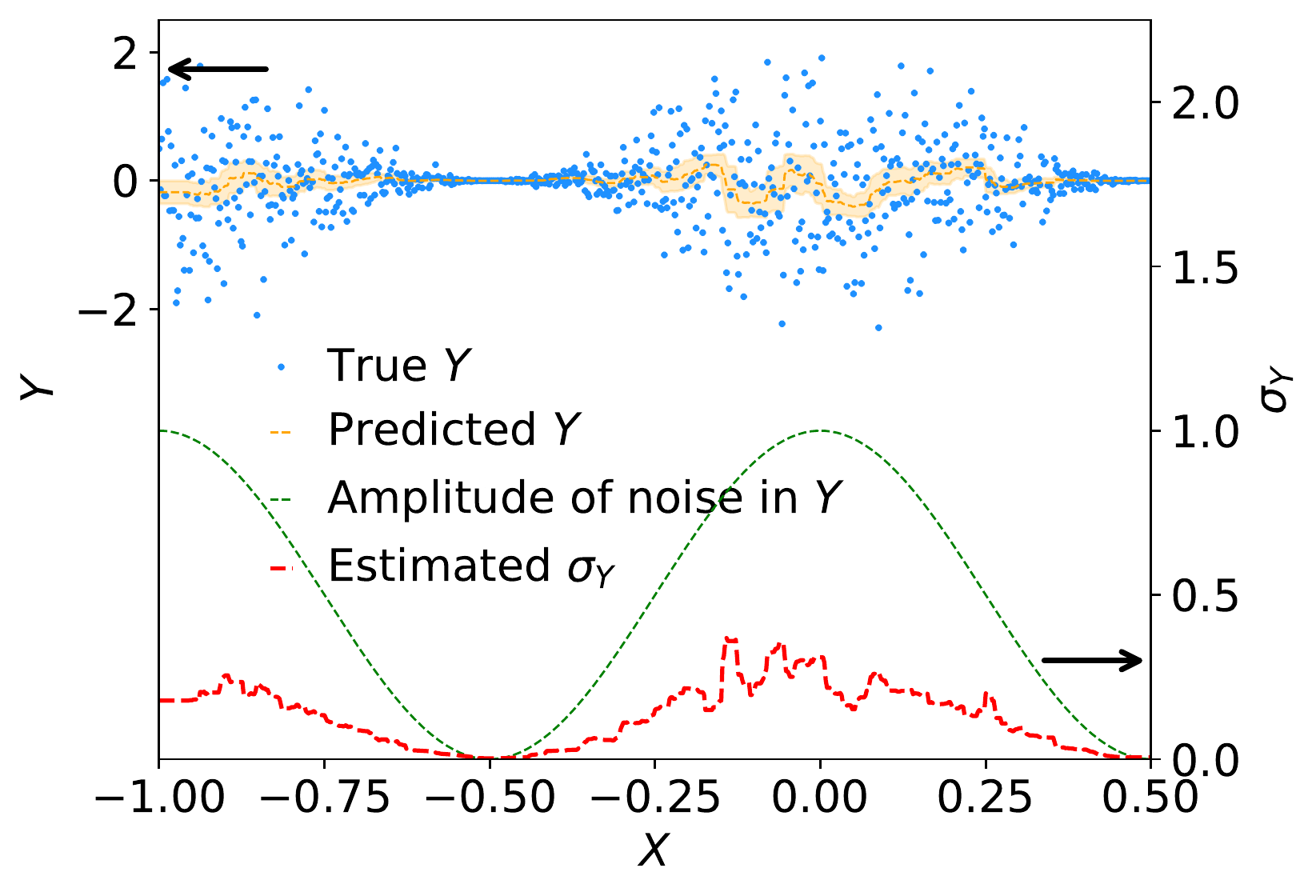}
  \caption{}
  \label{fig8:sub1}
\end{subfigure}\\
\begin{subfigure}{.47\textwidth}
  \includegraphics[width=\linewidth]{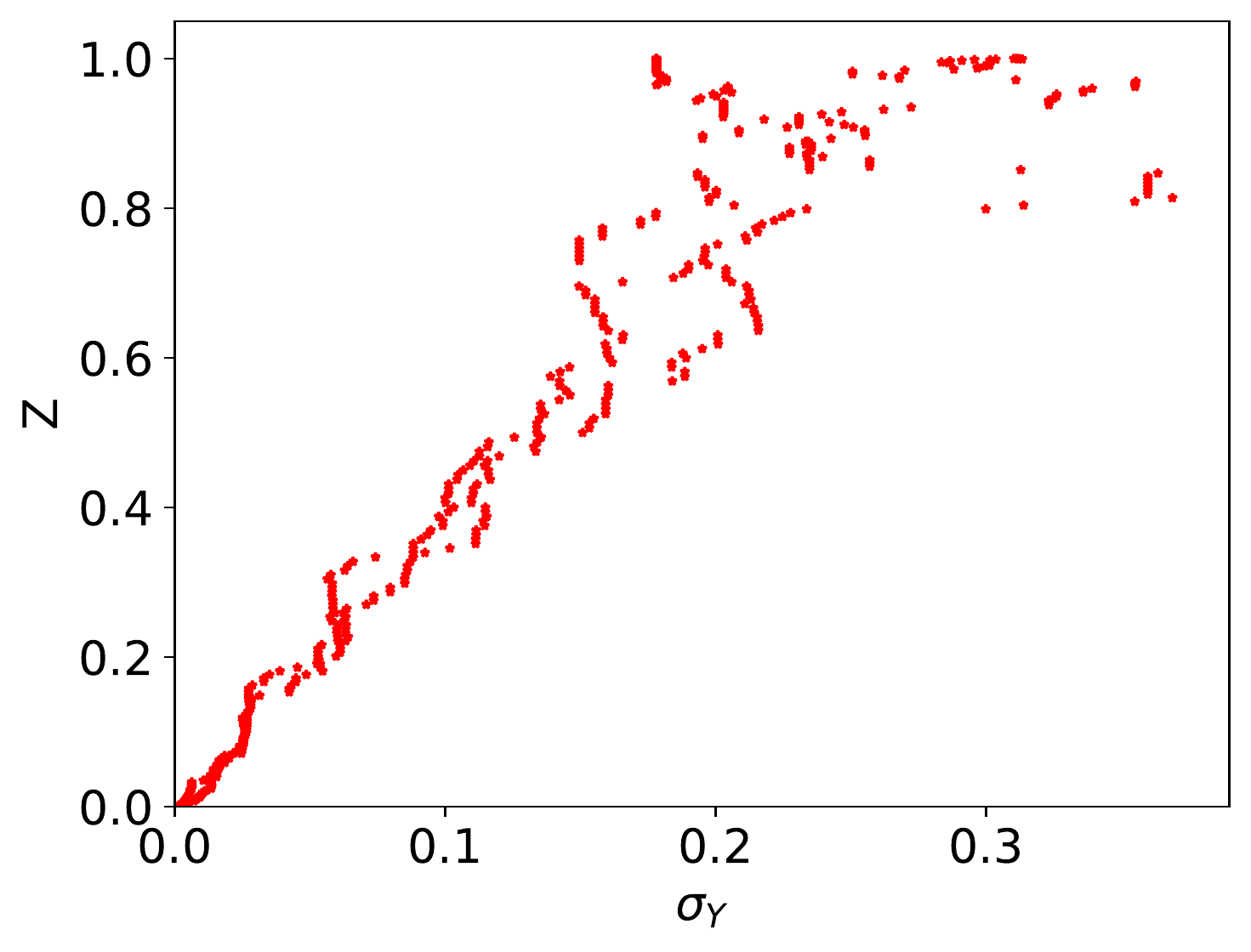}
  \caption{}
  \label{fig8:sub2}
\end{subfigure}
\begin{subfigure}{.49\textwidth}
  \includegraphics[width=\linewidth]{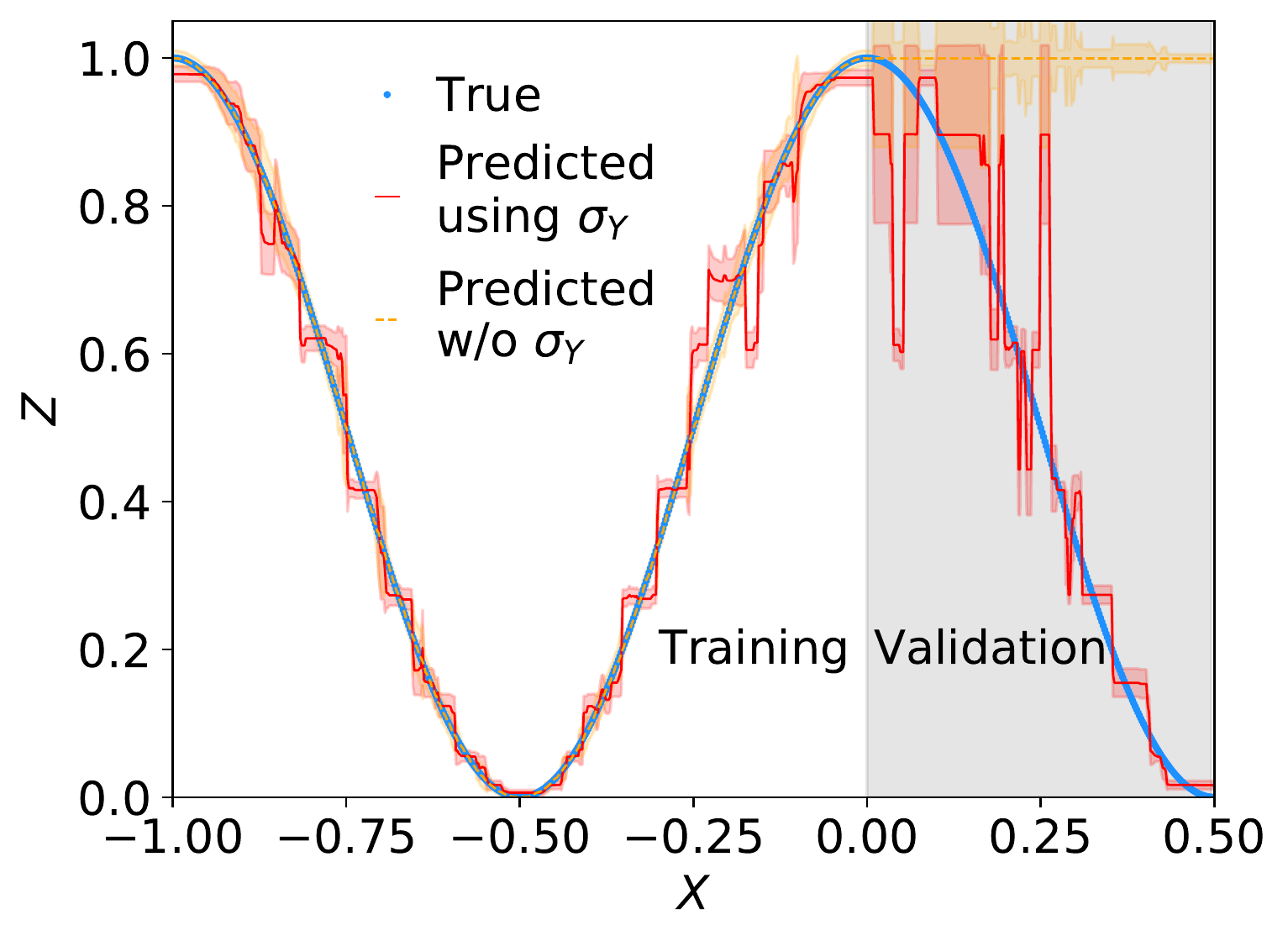}
  \caption{}
  \label{fig8:sub3}
\end{subfigure}
\end{figure}

Following the prescription of our study of missing data, we focus on the case of training on one period ($-1 < X < 0$) of data for $Z(X)$, shown in Fig. \ref{fig8}. We find the hyperparameters that maximize $R^2$ in $Z$-predictions calculated following the blocking cross-validation \cite{ref51}. The optimal value of $min\_samples\_leaf$ was 18, leading to the estimate of the error in predictions, $\sigma_Y$, being proportional to the noise in the underlying data, with a proportionality factor of $\frac{1}{\sqrt{18}}$ (Fig. \ref{fig8:sub1}, \ref{fig8:sub2}). The model with the tuned hyperparameters was trained and then used to predict $Z$ at $X > 0$. These predictions were compared against the true values, giving validation $R^2$ of 0.89 (red curve) and feature importance of $\sigma_Y$ of 0.93. It should be noted, however, that $Z$-predictions on validation (Fig. \ref{fig8:sub3}) could have been better - the plateau of $Z$-predictions extends up to $X = 0.21$, and the minimum value of $Z$ is 0.017 instead of 0. This stems from the noisy estimate of uncertainty in the random forest predictions of $Y$, meaning that the regressor chooses to learn partly from $X \rightarrow Z$. Without using $\sigma_Y$, the model mostly learns $X \rightarrow Z$, leading to approximately constant prediction of $Z$ (orange curve).

\begin{figure}[h]
\centering
\begin{subfigure}{.47\textwidth}
  \includegraphics[width=\linewidth]{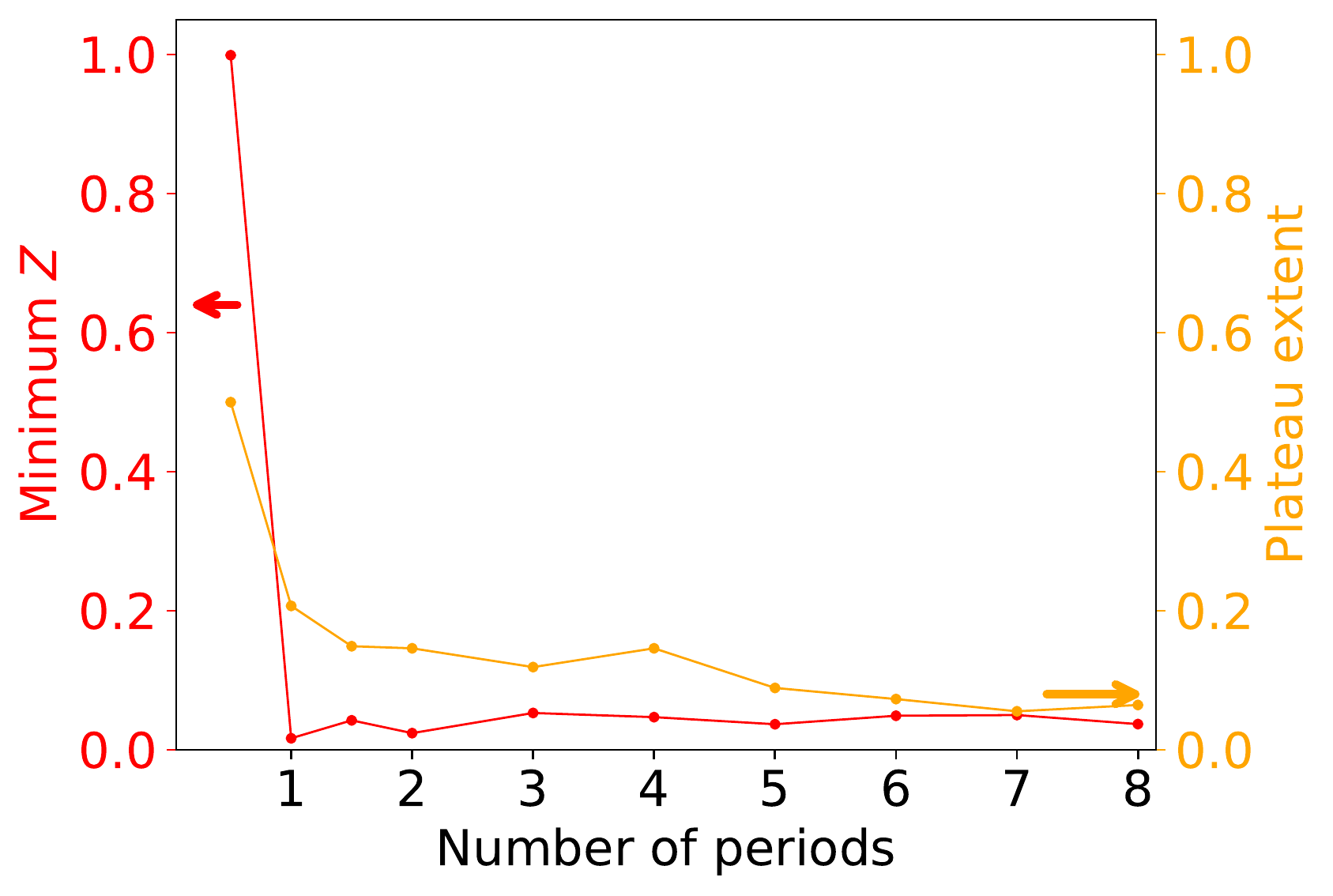}
  \caption{}
  \label{fig9:sub1}
\end{subfigure}%
\begin{subfigure}{.5\textwidth}
  \includegraphics[width=\linewidth]{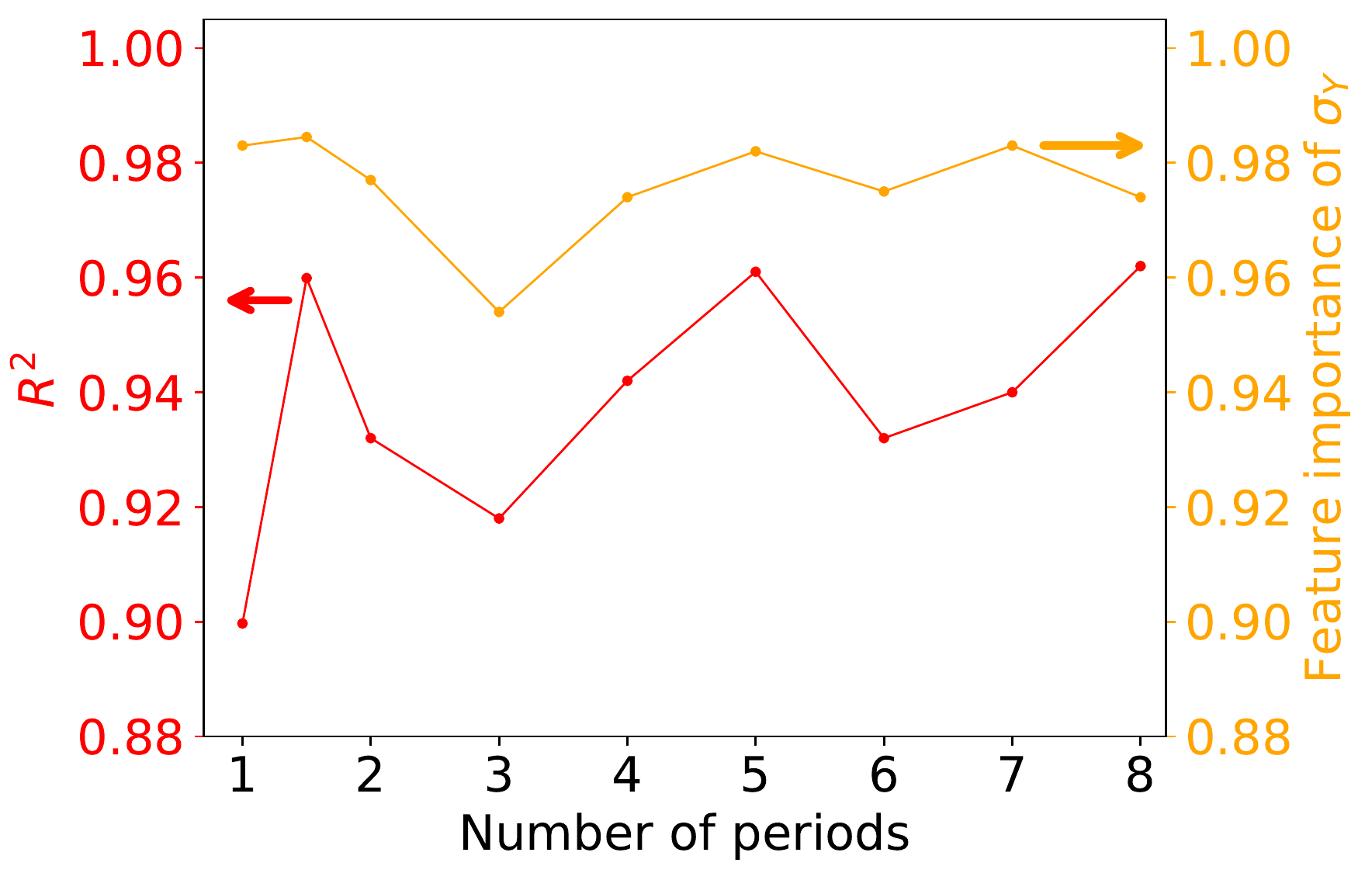}
  \caption{}
  \label{fig9:sub2}
\end{subfigure}
\caption{Effect of the number of periods in training set on: (a) Value of $Z$ at the minimum (red) and extent of plateau in $Z$-predictions (orange); (b) $R^2$ for validation data (red) and feature importance of $\sigma_Y$ (orange). Note that at half period, feature importance of $\sigma_Y$ and $R^2$ are too low ($0.003$ and $-1.98$ respectively) and therefore not shown in the plot.}
\label{fig9}
\end{figure}

To study the relative importance of the $X \rightarrow Z$ versus the $X \rightarrow \sigma_Y \rightarrow Z$ approach we now increase the lower bound $B$ of the range $B < X < 0$ of training data and therefore the number of periods of training data at $X < 0$ to increase the amount of data to learn the linear $\sigma_Y \rightarrow Z$ relationship. The results can be seen in Fig. \ref{fig9}. With more periods in training set, more information is available for learning the linear $\sigma_Y \rightarrow Z$ relationship, and learning $X \rightarrow Z$ becomes less favourable. This leads to increase in feature importance of $\sigma_Y$, consequently improving $R^2$, minimum $Z$, and the extent of the plateau on validation. Generally, as little as one period in the training data already gives a real-life benefit from using the uncertainty.

The amplitude of the noise, as long as it is above zero, does not affect the quality of predictions. This is expected, since random forest is scale-invariant. Tests presented so far have been for Gaussian distributed noise. Therefore, the algorithm was tested for other noise distributions including Cauchy, uniform and exponential. For all of these the algorithm delivered predictions with a similar level of accuracy.

\subsection{Extrapolation using both intermediate target variable and uncertainty}
\label{s34}
We have demonstrated that the multilayer regressor can utilize either an intermediate target variable or its uncertainty for extrapolation. If both the intermediate target variable and its uncertainty contain information about the final target variable but are individually noisy, it is possible to combine them to use the less noisy average to help extrapolate the final target variable. This would reduce the mean squared error in predictions of the final target variable by a factor of up to 2. Shot noise \cite{ref49} is one real-life example of where a variable and its uncertainty are related and so both contain information that we can exploit.

In order to validate the algorithm's ability to combine an intermediate target variable and its uncertainty for extrapolation, we construct a paradigmatic dataset with $X$, $Y$, and $Z$ columns. In this dataset, $X$ is the feature column, and $Y \sim \mathcal{N}(Z(X), b^2\vert Z(X)\vert^2)$, where $b$ is a positive real number. As before, we adopt $Z(X) = $ cos$^2(\pi X)$, with one period ($-1 < X < 0$) of $Z(X)$ for training and $Z$-values missing for $X > 0$ for validation.

To enable learning of $Z(X)$ from a linear combination of $Y$ and $\sigma_Y$, we first standardize $Y$ and $\sigma_Y$ computed by the first random forest model $X \rightarrow Y, \sigma_Y$. Then we perform a rotation of $Y$ and $\sigma_Y$ in the $Y-\sigma_Y$ plane by angle $\theta$: 
\begin{equation}
\begin{pmatrix}
           Y' \\
           \sigma_Y' 
\end{pmatrix} = \begin{pmatrix}
           \text{cos}(\theta),  - \text{sin}(\theta) \\
           \text{sin}(\theta),  \text{cos}(\theta)
\end{pmatrix} \begin{pmatrix}
           Y \\
           \sigma_Y 
\end{pmatrix}
\label{eqn3}
\end{equation}
After that, the two rotated components, $Y'$ and $\sigma_Y'$, are used as inputs by the second random forest model to learn $Z$.

\begin{figure}
\centering
\includegraphics[width=0.67\linewidth]{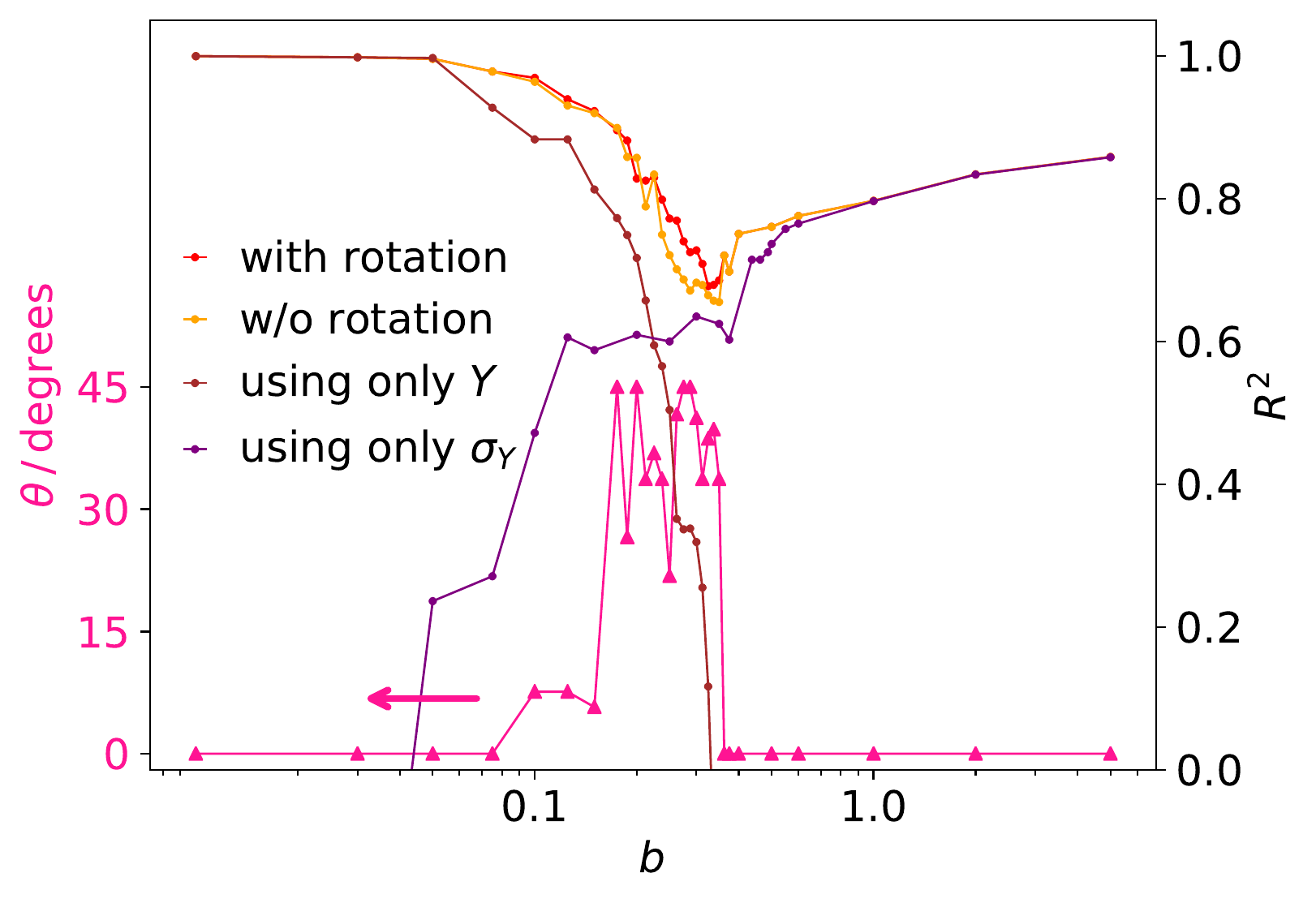}
\caption{$R^2$ for validation data at different $b$ using rotation in $Y-\sigma_Y$ plane (red), without rotation in $Y-\sigma_Y$ plane (orange), using only $Y$ (brown) and using only $\sigma_Y$ (purple). The pink curve shows the optimal angle of rotation $\theta$ in $Y-\sigma_Y$ plane.}
\label{fig10}
\end{figure}

First, working on the training set at $X < 0$, we find the hyperparameters that maximize $R^2$ in $Z$-predictions following the blocking cross-validation \cite{ref51}. The model with the tuned hyperparameters and the optimal angle $\theta$ of rotation in the $Y-\sigma_Y$ plane was trained and then used to predict $Z$ at $X > 0$. The predictions are compared against the true values for several values of $b$, and the results are presented in Fig. \ref{fig10}.

The $R^2$ values obtained when using rotation (red curve) are slightly better than the $R^2$ values obtained without using rotation (orange curve) due to the combination of $Y$ and $\sigma_Y$ being less noisy than the individual quantities. We also compute the $R^2$ values obtained when the second random forest model uses only $Y$ (brown curve) and only $\sigma_Y$ (purple curve). It is clear that for the small and large values of $b$, which correspond to the limits discussed in Sections \ref{s32} and \ref{s33}, no rotation is necessary due to the fact that $Z$ is predominantly learnt from $Y$ and $\sigma_Y$ respectively. Both red and orange curves approach the brown curve in the former limit and the purple curve in the latter limit from above.

The noise in $\sigma_Y$ is smaller than the noise in $Y$ by a factor of $\approx \frac{\sqrt{2}}{b}$, therefore $\theta = 0$ at $b > 0.7$, i.e.\,\,it is easier to use only $\sigma_Y$ for predictions. At at the intermediate values of $b$ less than 0.7, however, $\theta$ peaks at 45 degrees, meaning that rotation combines the information from $Y$ and $\sigma_Y$ in equal proportions to predict $Z$. Using this rotation marginally improves the predictions (red curve) compared to not using any rotation (orange curve). This improvement comes from averaging the noise in $Y$ and $\sigma_Y$, which would otherwise exacerbate predictions of $Z$ in regions where $Y$ and $\sigma_Y$ are particularly noisy. 
\section{Application to real-world physical examples}
\label{s4}
Having set up and validated the machine learning algorithm for extracting information from uncertainty, both directly and alongside the expected value, we are well-positioned to test these two capabilities of the formalism respectively on two real-life physical examples: dielectric crystal phase transitions (Section \ref{s41}) and single-particle diffraction of droplets (Section \ref{s42}). In each case, we take experimental data from the literature and split it into two tranches. We then train the model on the first tranche and validate against the second tranche to replicate a real-life blind prediction against a future experiment.

A general characteristic of a system that machine learning from uncertainty will benefit is non-monotonic $X \rightarrow Z$ behaviour coupled with a noisy intermediate variable $Y$. An excellent example is a material that undergoes multiple phase transitions as tuning parameter ($X$) increases. Fluctuations in the system's order parameter ($Y$) will always be elevated near to each phase transition \cite{ref14}, and so will the energy associated with fluctuations, and therefore heat capacity ($Z$). Precise measurements of heat capacity are typically performed using differential scanning calorimetry \cite{ref17}, which is usually more costly than measuring order parameter (e.g.\,\,dielectric constant). It is therefore attractive to learn $X \rightarrow \sigma_Y$ and then $\sigma_Y \rightarrow Z$, i.e.\,\,a map from order parameter fluctuations to heat capacity, in order to extrapolate the latter. In Section \ref{s41} we apply exactly this approach to a dielectric crystal phase transition.

Systems where combining the value of the intermediate quantity with its uncertainty to achieve statistical averaging are often characterized by counting with shot noise \cite{ref49}. An example is single-particle diffraction, which we study in Section \ref{s42}. Here, as the diffraction angle ($X$) varies, the particle count ($Y$) exhibits noise ($\sigma_Y$) that is dependent on $Y$. The combination of the particle count and its uncertainty can be used to predict the ground truth diffraction pattern ($Z$), which is obscured in the regions where the particle count itself is finite and therefore noisy. 
\subsection{PbZr$_{0.7}$Sn$_{0.3}$O$_{3}$ crystal phase transitions}
\label{s41}
\begin{figure}
\centering
\begin{subfigure}{.49\textwidth}
  \includegraphics[width=\linewidth]{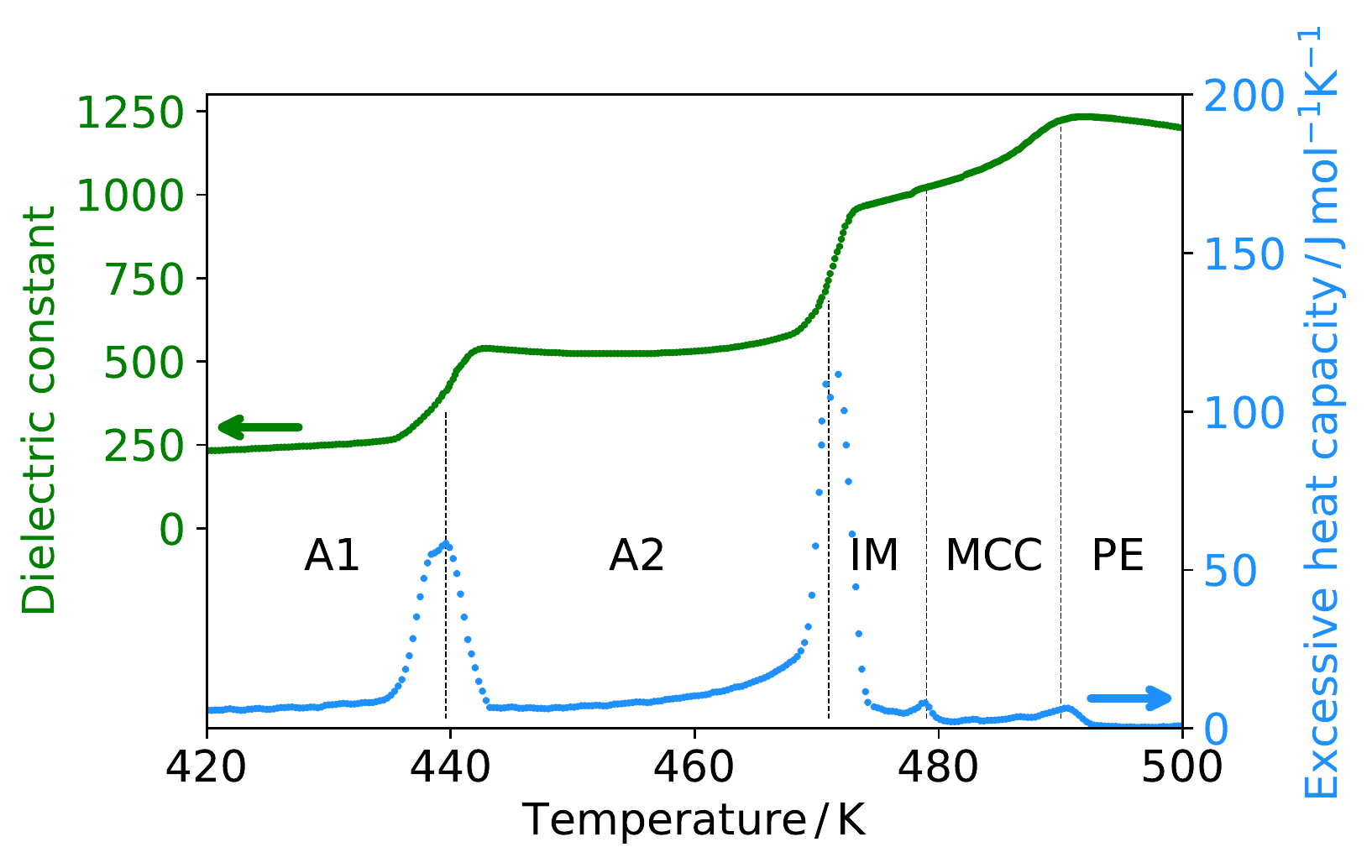}
  \caption{}
  \label{fig11:sub1}
\end{subfigure}%
\begin{subfigure}{.49\textwidth}
  \includegraphics[width=0.895\linewidth]{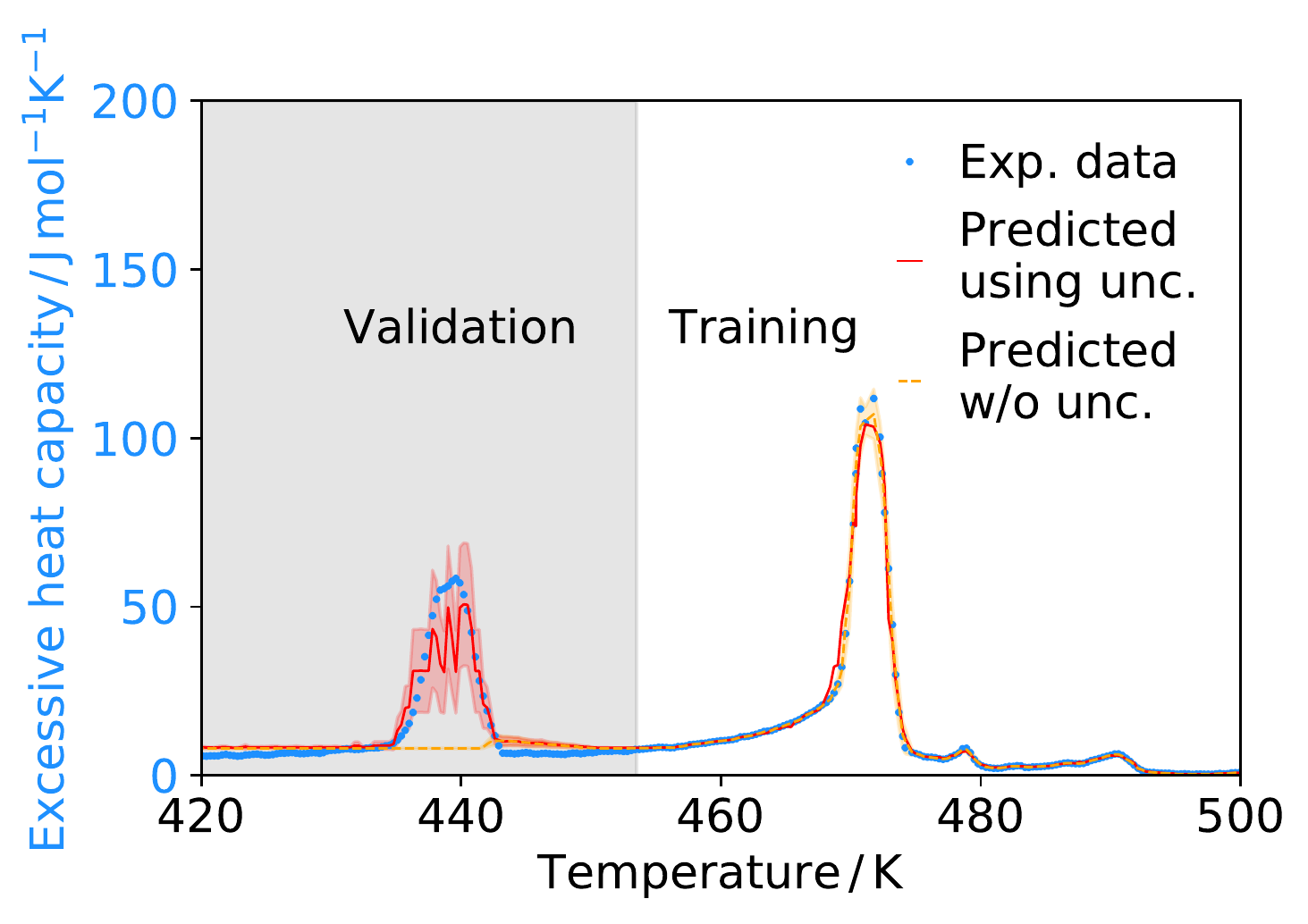}
  \caption{}
  \label{fig11:sub2}
\end{subfigure}
\caption{(a) Experimental data on dielectric constant (green, primary $y$-axis) and excessive heat capacity (blue, secondary $y$-axis) against temperature. The vertical black dotted lines denote the phase transitions between the A1, A2, IM, MCC and PE phases. Reproduced from Fig. 5 in Ref. \cite{ref15}.  (b) Predictions of the phase transition at $\sim 440$ K. The plot includes predictions using uncertainty (red, error region shaded, $R^2 = 0.85$ on validation) and without uncertainty (orange, error region shaded, $R^2 = -0.04$ on validation). The grey shaded area is the validation region.}
\label{fig11}
\end{figure}

An excellent example of a system that undergoes multiple phase transitions is PbZr$_{0.7}$Sn$_{0.3}$O$_{3}$ -- a dielectric crystalline solid. As the temperature ($X$) increases, this crystal goes from antiferroelectric state (A1) to paraelectric state (PE) via intermediate states A2, IM, and MCC \cite{ref15} so passes through a total of four phase transitions. The order parameter ($Y$) is the dielectric constant. The experimental data \cite{ref15} available for heat capacity already accounts for the Debye contribution \cite{ref16}, leaving the excessive heat capacity ($Z$) associated with phase transitions, i.e.\,\,order parameter fluctuations. The plot of the available experimental data \cite{ref15} is shown in Fig. \ref{fig11:sub1}.

We focus on first-order phase transitions A1 $\leftrightarrow$ A2 and A2 $\leftrightarrow$ IM. This pair of first-order transitions are the most prominent in Fig. \ref{fig11:sub1}, whereas the other second-order transitions, despite having diverging fluctuations, have a narrow region of sharp increase in heat capacity that is not resolved in the available experimental data.

We train a multilayer regressor on the peak at $\sim 471$ K and validate on the peak at $\sim 440$ K. For training, the dielectric constant data is available at all temperatures, but the excessive heat capacity is only available to the right of the first peak, therefore provides no information about the phase transition at $\sim 440$ K. The algorithm delivers strong predictions for heat capacity when compared to unseen data at temperatures below $455$ K with $R^2 = 0.85$, correctly identifying the phase transition at $\sim 440$ K. The feature importance of uncertainty in dielectric constant is 0.87, showing the utility of uncertainty in understanding the phase transitions. If the machine learning is trained without being able to exploit uncertainty, we see the predictions completely miss the phase transition with $R^2 = -0.04$, confirming the importance of predicting and exploiting the uncertainty.
\subsection{Single-particle diffraction of droplets}
\label{s42}
Having successfully applied our method to dielectric crystal phase transitions, we proceed to demonstrate its applicability to another phenomenon -- diffraction of droplets through a double slit, taking data from Ref. \cite{ref50}. For a given angle ($X$) the count of droplets ($Y$) diffracted into that angle was measured and plotted on a histogram. The flow of droplets is low, making accumulation of data for the diffraction pattern time-consuming. We therefore turn to machine learning to take available data and estimate the diffraction pattern, expecting it to be that of a double slit. The noise in the count ($\sigma_Y$) is expected to follow a Poisson distribution \cite{ref12}, i.e.\,\,to depend on the expected value of the count. Therefore, it is possible to use both $Y$ and $\sigma_Y$ self-averaging the statistical uncertainty in each, delivering a more precise estimate for the analytical ground truth amplitude ($Z$), despite the finite number of droplets in the experiment. The determination of the ground truth amplitude is useful for investigating the properties of the particles source (e.g.\,\,particle energy) if at some angles the particle count is noisy or low.
\begin{figure}
\centering
\begin{subfigure}{.49\textwidth}
  \includegraphics[width=\linewidth]{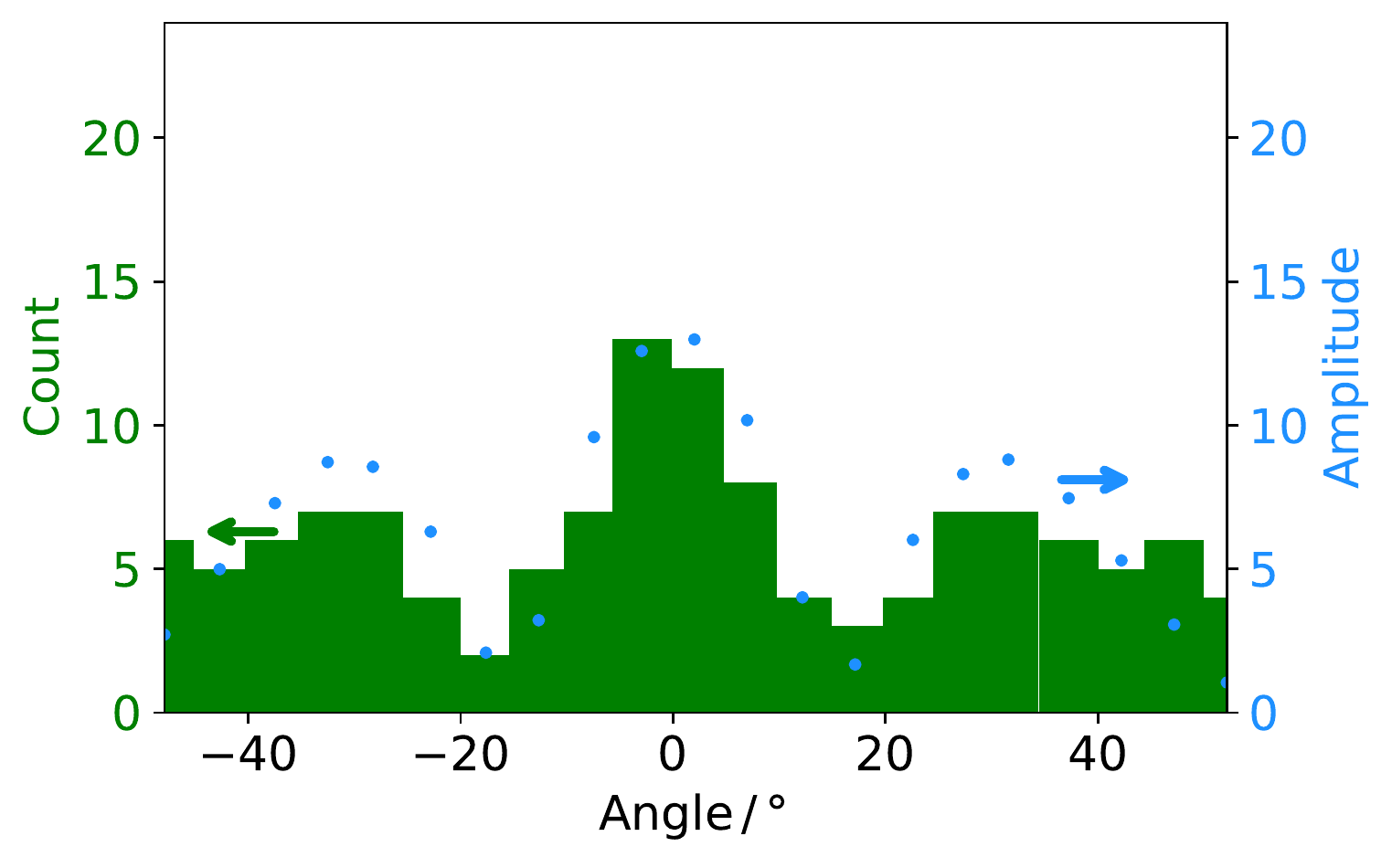}
  \caption{}
  \label{fig12:sub1}
\end{subfigure}%
\begin{subfigure}{.49\textwidth}
  \includegraphics[width=0.92\linewidth]{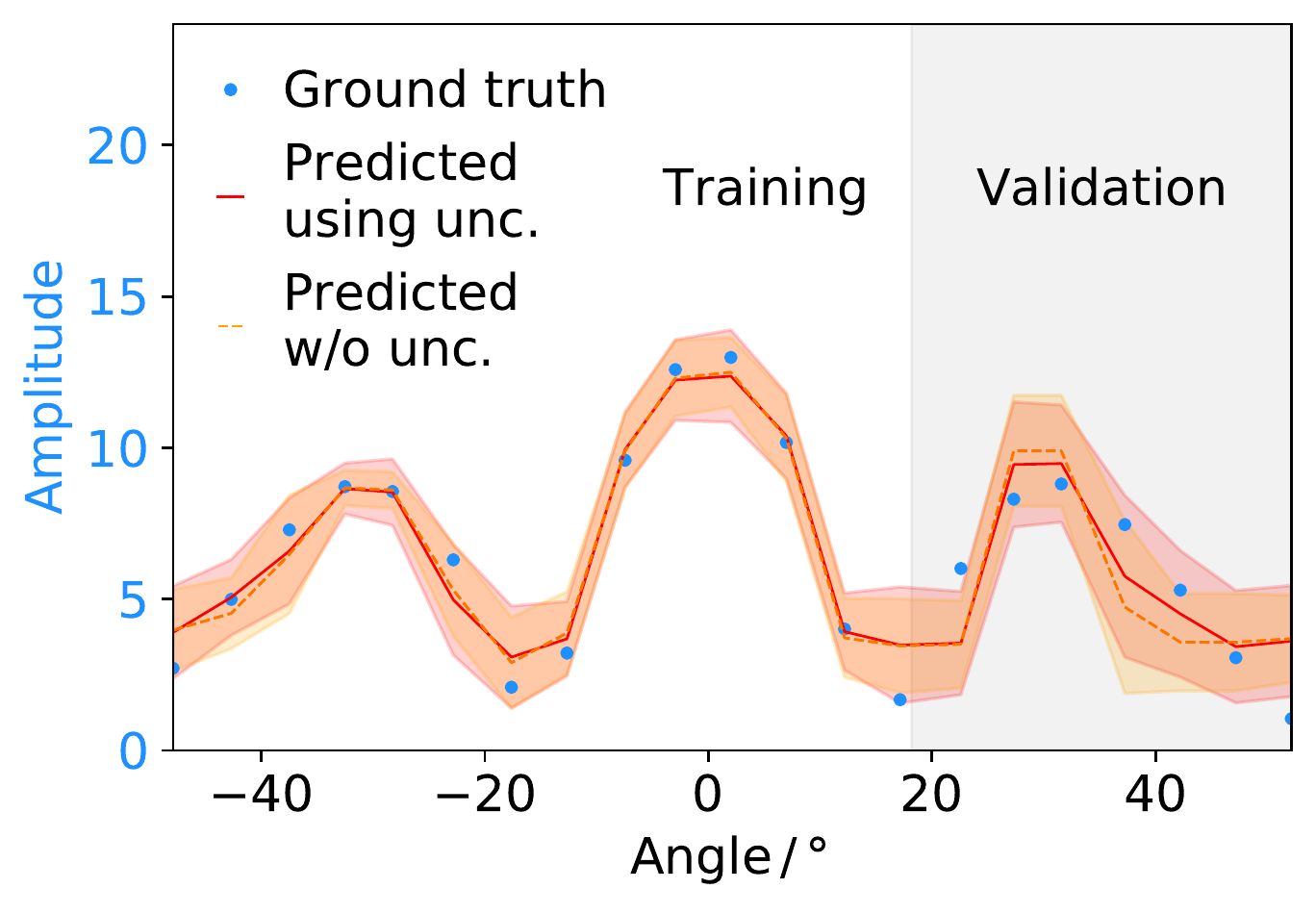}
  \caption{}
  \label{fig12:sub2}
\end{subfigure}
\caption{(a) Data for droplet count (green, primary $y$-axis) and ground truth amplitude (blue, secondary $y$-axis) with angle, reproduced from Fig. 3 in Ref. \cite{ref50}. (b) Predictions of the amplitude for the peak at $\sim 30\degree$ with the validated region shaded grey. The plot includes predictions using uncertainty (red, error region shaded, $R^2 = 0.63$ on validation) and without uncertainty (orange, error region shaded, $R^2 = 0.43$ on validation), and original data are blue points. The grey shaded area is the validation region.}
\label{fig12}
\end{figure}

Experimental data for diffraction of 75 particles was taken from Ref. \cite{ref50} and is shown in Fig. \ref{fig12:sub1}. We train multilayer regressor on the two peaks at $\sim -30\degree$ and $\sim 0\degree$ and validate on the peak at $\sim 30\degree$. For training, the count is available at all angles, but the ground truth amplitude is only available at the peaks at $\sim -30\degree$ and $\sim 0\degree$. Such a situation may arise when the ground truth amplitude is unknown at some angles due to the complex nature of the particles source and/or the aperture. The results can be seen in Fig. \ref{fig12:sub2}. The model achieves $R^2 = 0.63$ on validation. Without using the uncertainty, the value of $R^2$ on the validation set is $0.43$, confirming the significant benefit of the use of uncertainty to improve extrapolation. The mean squared error in predictions is reduced by a factor of $\frac{1 - 0.43}{1 - 0.63} = 1.54 < 2$, in agreement with the expected improvement when using uncertainty discussed in Section \ref{s34}.
\section{Conclusion}
\label{s5}
We developed, implemented, and validated a machine learning framework which, given an input feature $X$, calculates uncertainty in target variable $Y$, $\sigma_Y$, and uses $Y$ and/or $\sigma_Y$ to predict another target variable $Z$. Two successive interpolations $X \rightarrow Y, \sigma_Y$ and $Y, \sigma_Y \rightarrow Z$ enable the difficult extrapolation $X \rightarrow Z$. Tests on paradigmatic datasets show two significant advantages: firstly the exploitation of information only in $\sigma_Y$, and secondly reduction of noise by averaging $Y$ and $\sigma_Y$ when they are proportional.

To showcase the method it was applied to two experimental datasets. For the first dataset on dielectric crystal, given the temperature ($X$) range and the order parameter ($Y$) values and exploiting the uncertainty $\sigma_Y$ in $Y$-predictions, heat capacity ($Z$) was extrapolated with respect to temperature. The method quantitatively predicted the phase transition completely missed by standard machine learning methods. For the second dataset, single-particle diffraction of droplets, given the angle ($X$) range and the particle count ($Y$) values and exploiting the $Y$-values together with the uncertainty $\sigma_Y$ in $Y$-predictions, the ground truth amplitude ($Z$) was extrapolated with respect to angle. Our method that combines $Y$ and $\sigma_Y$ improves extrapolation in the region with noisy $Y$-values, reducing the mean squared error by a factor of $\sim 2$. This demonstrates the importance of uncertainty as a source of information in its own right, to improve the predictive power of machine learning methods for physical phenomena.

Furthermore, the method can operate on any number of input features and target variables and the generic algorithm can be applied in many different situations. This endorses the method's applicability to more complex physical systems, e.g.\,\,concrete or atomic junctions. For concrete, the input feature $X$ would be the position within the image of the material's microstructure. The intermediate target variable $Y$ would be the size/contrast of the aggregate, and the uncertainty in it, $\sigma_Y$, is linked to mechanical properties of the material ($Z$), such as strength \cite{ref11}. For atomic junctions, the input feature $X$ would be the shape/structure of the junction. The intermediate target variable $Y$ would be the count of electrons passing through the junction. This count has shot noise, hence the combination of $Y$ and $\sigma_Y$ can be used to improve predictions of properties linked to electron count, such as conductivity ($Z$) \cite{ref58}.

The algorithm has potential applications in areas beyond physics as well. One of these areas is financial markets, where higher uncertainty in predictions of future stock price movement leads to investors being less likely to buy or sell it, i.e.\,\,to decrease in its trading volume \cite{ref34}. Another example is cancer, which is known to cause genetic chaos \cite{ref47}. The information extracted from this chaos can be used for early cancer detection.
\section*{Acknowledgments}

The authors acknowledge the financial support of the Engineering and Physical Sciences Research Council, 
Harding Distinguished Postgraduate Scholars Programme Leverage Scheme, and the Royal Society 
(Grant No. URF{\textbackslash}R{\textbackslash}201002). There is Open Access to this paper at \texttt{https://www.openaccess.cam.ac.uk}. For the purpose of open access, the author has applied a Creative Commons Attribution (CC BY) licence to any Author Accepted Manuscript version arising.

\section*{Statements and Declarations}
Gareth Conduit is a Director of materials machine learning company Intellegens. The authors have no other financial interests or personal relationships that could have appeared to influence the work reported in this paper.
\section*{CRediT authorship contribution statement}
\textbf{Bahdan Zviazhynski}: Methodology, Software, Formal analysis, Data curation, Writing -- original draft. \textbf{Gareth Conduit}: Conceptualisation, Methodology, Writing -- review \& editing, Supervision.
\bibliographystyle{mystyle}
\bibliography{bibliography}

\end{document}